\documentclass{article}

\usepackage[utf8]{inputenc}
\usepackage[T1]{fontenc}
\usepackage{hyperref}
\usepackage{url}
\usepackage{booktabs}
\usepackage{amsfonts}
\usepackage{nicefrac}
\usepackage{xcolor}
\usepackage{graphicx}
\usepackage{amsmath}
\usepackage{amssymb}
\usepackage{algorithm}
\usepackage{algorithmic}
\usepackage{subcaption}
\usepackage[margin=1in]{geometry}
\usepackage{natbib}
\usepackage{multirow}
\usepackage{float}

\DeclareMathOperator*{\argmin}{arg\,min}

\newcommand{\phivec}{\boldsymbol{\phi}}
\newcommand{\Phiglobal}{\Phi}
\newcommand{\alphavec}{\boldsymbol{\alpha}}
\newcommand{\betavec}{\boldsymbol{\beta}}
\newcommand{\hvec}{\mathbf{h}}
\newcommand{\E}{\mathbf{E}}
\newcommand{\R}{\mathbb{R}}

\title{Differentiable Symbolic Planning: A Neural Architecture for\\Constraint Reasoning with Learned Feasibility}

\author{
  Venkatakrishna Reddy Oruganti\\
  Sithara Inc.\\
  Plano, Texas, USA\\
  \texttt{venkatakrishnareddy.oruganti@sitharaai.com}
}

\date{February 2026}

\begin{document}

\maketitle

\begin{abstract}
Neural networks excel at pattern recognition but struggle with constraint reasoning---determining whether configurations satisfy logical or physical constraints. We introduce \textbf{Differentiable Symbolic Planning (DSP)}, a neural architecture that performs discrete symbolic reasoning while remaining fully differentiable. DSP maintains a \textbf{feasibility channel} ($\phi$) that tracks constraint satisfaction evidence at each node, aggregates this into a \textbf{global feasibility signal} ($\Phi$) through learned rule-weighted combination, and uses \textbf{sparsemax attention} to achieve exact-zero discrete rule selection. We integrate DSP into a \textbf{Universal Cognitive Kernel (UCK)} that combines graph attention with iterative constraint propagation. Evaluated on three constraint reasoning benchmarks---graph reachability, Boolean satisfiability, and planning feasibility---UCK+DSP achieves 97.4\% accuracy on planning under 4$\times$ size generalization (vs.\ 59.7\% for ablated baselines), 96.4\% on SAT under 2$\times$ generalization, and maintains balanced performance on both positive and negative classes where standard neural approaches collapse. Ablation studies reveal that global $\phi$ aggregation is critical: removing it causes accuracy to drop from 98\% to 64\%. The learned $\phi$ signal exhibits interpretable semantics, with values of $+18$ for feasible cases and $-13$ for infeasible cases emerging without supervision.
\end{abstract}


%==============================================================================
\section{Introduction}
%==============================================================================

Constraint reasoning---determining whether a configuration satisfies a set of logical, physical, or relational constraints---is fundamental to planning, verification, and decision-making. Given a graph representing a state space, can an agent reach a goal? Given a Boolean formula, is it satisfiable? Given a gridworld with obstacles, does a feasible path exist? These problems share a common structure: binary classification over structured inputs where the answer depends on global constraint satisfaction.

Neural networks have achieved remarkable success in pattern recognition, yet they exhibit systematic failures on constraint reasoning tasks. The core difficulty is \textbf{class collapse under distribution shift}: when tested on problems larger than those seen during training, neural models achieve high accuracy on one class (typically the negative/infeasible class) while catastrophically failing on the other. A model reporting 90\% overall accuracy may be correctly rejecting 99\% of infeasible cases while accepting only 20\% of feasible ones---operationally useless for any application requiring balanced decisions.

We identify three architectural gaps in existing neural approaches:

\begin{enumerate}
    \item \textbf{No explicit feasibility tracking.} Standard graph neural networks (GNNs) and transformers lack dedicated state variables for accumulating constraint satisfaction evidence across reasoning steps.
    
    \item \textbf{Dense attention prevents discrete reasoning.} Softmax attention distributes weight across all elements, preventing the discrete rule selection necessary for constraint propagation. When rule 3 should fire and rules 1, 2, 4 should not, softmax still assigns them nonzero weight.
    
    \item \textbf{No global aggregation mechanism.} Local node representations alone cannot capture whether a \emph{global} constraint (e.g., path existence) is satisfied. Existing architectures rely on pooling operations that lose the structured reasoning signal.
\end{enumerate}

We address these gaps with \textbf{Differentiable Symbolic Planning (DSP)}, a neural module with three key innovations:

\textbf{Feasibility Channel ($\phi$).} Each node maintains a scalar $\phi$ value that accumulates evidence for or against constraint satisfaction. This channel persists across reasoning steps and directly informs the final prediction.

\textbf{Global Feasibility Aggregation ($\Phi$).} Local $\phi$ values are combined through learned, rule-weighted aggregation into a single global signal. This mechanism is critical---ablation shows removing it causes accuracy to collapse from 98\% to 64\%.

\textbf{Sparsemax Attention.} We replace softmax with sparsemax projections that produce exact zeros, enabling discrete rule selection while maintaining differentiability. This prevents interference between rules and yields 6$\times$ lower variance than softmax.

We integrate DSP into a \textbf{Universal Cognitive Kernel (UCK)} that alternates graph attention (for local message passing) with DSP updates (for constraint reasoning). The system processes graph-structured inputs through $T$ rollout steps, with $\phi$ and $\Phi$ evolving to reflect accumulated constraint evidence.

\paragraph{Contributions.}
\begin{enumerate}
    \item We introduce DSP, a differentiable module for constraint reasoning with explicit feasibility tracking, global aggregation, and sparse discrete attention.
    
    \item We demonstrate state-of-the-art performance on three benchmarks: planning feasibility (97.4\% under 4$\times$ size generalization), Boolean satisfiability (96.4\% under 2$\times$ generalization), and graph reachability (82.7\% under 2.5$\times$ generalization).
    
    \item We provide extensive ablations proving the necessity of each component: removing global $\phi$ drops accuracy by 34 points; replacing sparsemax with softmax drops accuracy by 26 points.
    
    \item We show that the learned $\phi$ signal exhibits interpretable semantics: feasible cases produce $\phi \approx +18$, infeasible cases produce $\phi \approx -13$, with 31-point separation emerging purely from training signal.
\end{enumerate}

%==============================================================================
\section{Related Work}
%==============================================================================

\paragraph{Graph Neural Networks for Reasoning.}
GNNs have been applied to combinatorial problems including SAT \citep{selsam2019learning}, planning \citep{toyer2018action}, and theorem proving \citep{paliwal2020graph}. However, these approaches lack explicit mechanisms for tracking constraint satisfaction and exhibit class collapse under size generalization. NeuroSAT \citep{selsam2019learning} pioneered neural SAT solving but relies on recurrent message passing without feasibility channels or sparse attention. Critically, \emph{NeuroSAT does not maintain a global feasibility state}---it relies on pooling over final node embeddings, which loses structured constraint information under distribution shift.

\paragraph{Neuro-Symbolic Systems.}
Hybrid approaches combine neural perception with symbolic reasoning \citep{garcez2019neural, mao2019neuro}. These often require hand-crafted symbolic components or domain-specific languages. DSP learns symbolic-like rules end-to-end while maintaining full differentiability, requiring no domain-specific engineering.

\paragraph{Sparse Attention.}
Sparsemax \citep{martins2016softmax} and entmax \citep{peters2019sparse} project attention weights onto the simplex, producing exact zeros. We show that sparse attention is critical for constraint reasoning---softmax attention causes systematic failures under generalization because it cannot achieve the discrete rule selection that constraint propagation requires.

\paragraph{Transformers for Structured Prediction.}
Transformers have been applied to graph problems \citep{ying2021transformers} and algorithmic reasoning \citep{velickovic2022clrs}. However, transformers lack dedicated constraint-tracking mechanisms and use dense attention that prevents discrete reasoning. \emph{Transformers are sequence predictors; DSP is a constraint reasoner}---these are complementary, not competing, architectural paradigms.

\paragraph{Planning with Neural Networks.}
Value Iteration Networks \citep{tamar2016value} and related architectures perform differentiable planning but focus on policy learning rather than feasibility classification. VIN answers ``what action should I take?'' while DSP answers ``does a valid plan exist?''---a prerequisite question that VIN cannot reliably answer under distribution shift.

%==============================================================================
\section{Problem Formulation}
%==============================================================================

We consider binary classification over graph-structured inputs where the label depends on global constraint satisfaction.

\paragraph{Input.} A graph $G = (V, E)$ with node features $\mathbf{X} \in \R^{N \times d}$ and adjacency matrix $\mathbf{A} \in \R^{N \times N}$. Task-specific designations include source/target nodes (for reachability/planning) or variable/clause structure (for SAT).

\paragraph{Output.} Binary prediction $y \in \{0, 1\}$ indicating whether the configuration satisfies the relevant constraint (reachable/unreachable, SAT/UNSAT, feasible/infeasible).

\paragraph{Evaluation Metrics.} Beyond overall accuracy, we measure:
\begin{itemize}
    \item \textbf{Per-class accuracy}: Separate accuracy on positive (feasible/SAT/reachable) and negative classes
    \item \textbf{Balance score}: $\min(\text{acc}_+, \text{acc}_-) / \max(\text{acc}_+, \text{acc}_-)$, where 1.0 indicates perfect balance
    \item \textbf{Size generalization}: Performance gap between training and test problem sizes
\end{itemize}

\paragraph{The Class Collapse Problem.}
Standard neural approaches optimize cross-entropy loss, which can be minimized by achieving high accuracy on the majority or easier class. Under distribution shift (larger problems at test time), this manifests as catastrophic failure on one class while maintaining accuracy on the other. A model achieving 99\% on infeasible cases but 20\% on feasible cases has 60\% overall accuracy but 0.20 balance---operationally useless.

%==============================================================================
\section{Method: Differentiable Symbolic Planning}
%==============================================================================

\subsection{Architecture Overview}

\begin{figure}[t]
    \centering
    \includegraphics[width=0.85\textwidth]{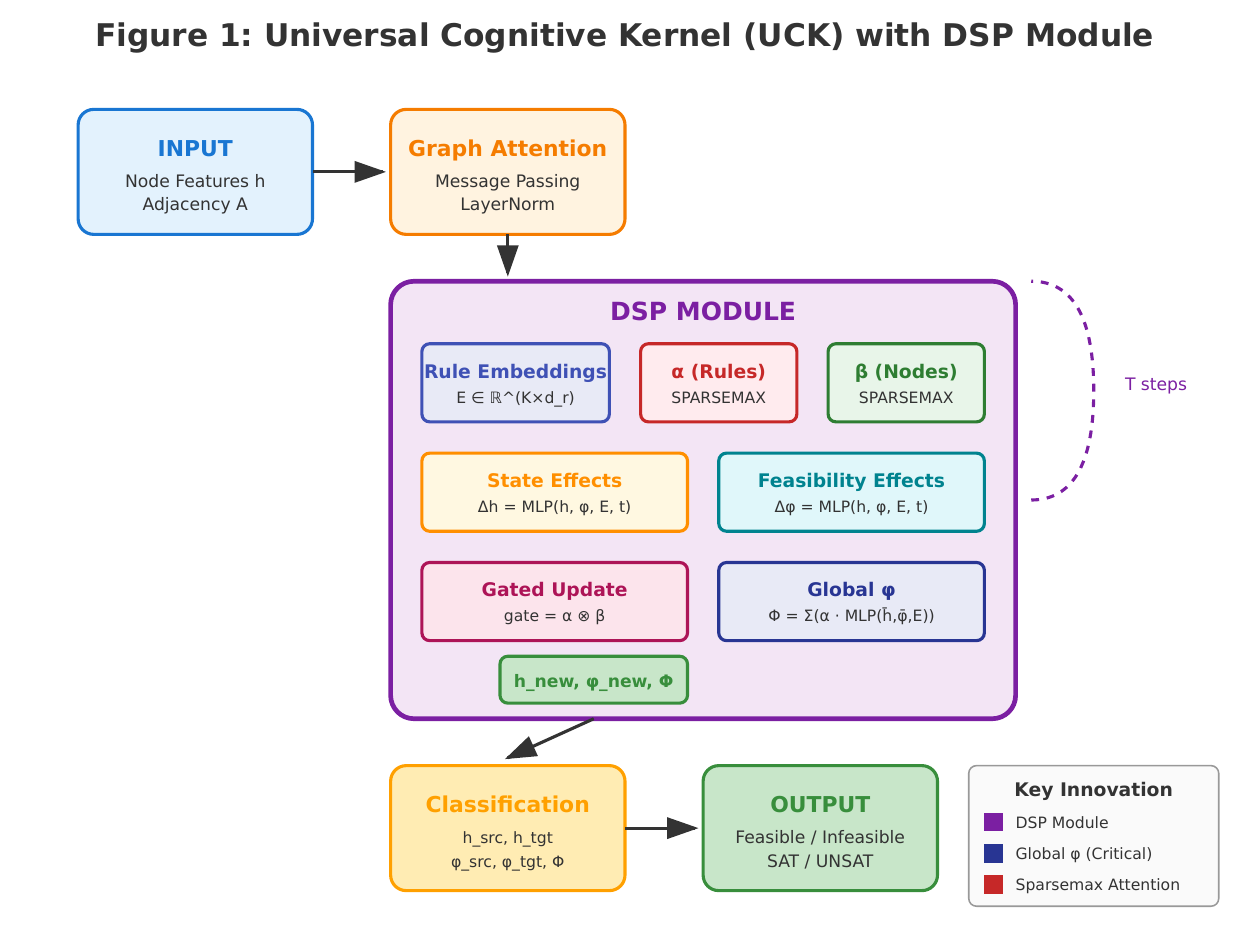}
    \caption{Universal Cognitive Kernel (UCK) with DSP Module. The system processes graph-structured inputs through $T$ rollout steps, alternating between graph attention and DSP updates. The DSP module maintains learnable rule embeddings, computes sparse rule activation ($\alpha$) and node selection ($\beta$) using sparsemax, and aggregates local feasibility into a global signal ($\Phi$).}
    \label{fig:architecture}
\end{figure}

The Universal Cognitive Kernel (UCK) with DSP processes inputs through $T$ rollout steps (Figure~\ref{fig:architecture}). Each step comprises:
\begin{enumerate}
    \item \textbf{Graph Attention}: Local message passing to propagate information along edges
    \item \textbf{DSP Update}: Rule-based reasoning with feasibility tracking
\end{enumerate}

The system maintains three state variables:
\begin{itemize}
    \item \textbf{Node states} $\hvec \in \R^{N \times d}$: Learned representations
    \item \textbf{Feasibility channel} $\phivec \in \R^{N \times 1}$: Per-node constraint satisfaction evidence
    \item \textbf{Global feasibility} $\Phiglobal \in \R$: Aggregated feasibility signal
\end{itemize}

\subsection{DSP Module}

\begin{figure}[t]
    \centering
    \includegraphics[width=0.9\textwidth]{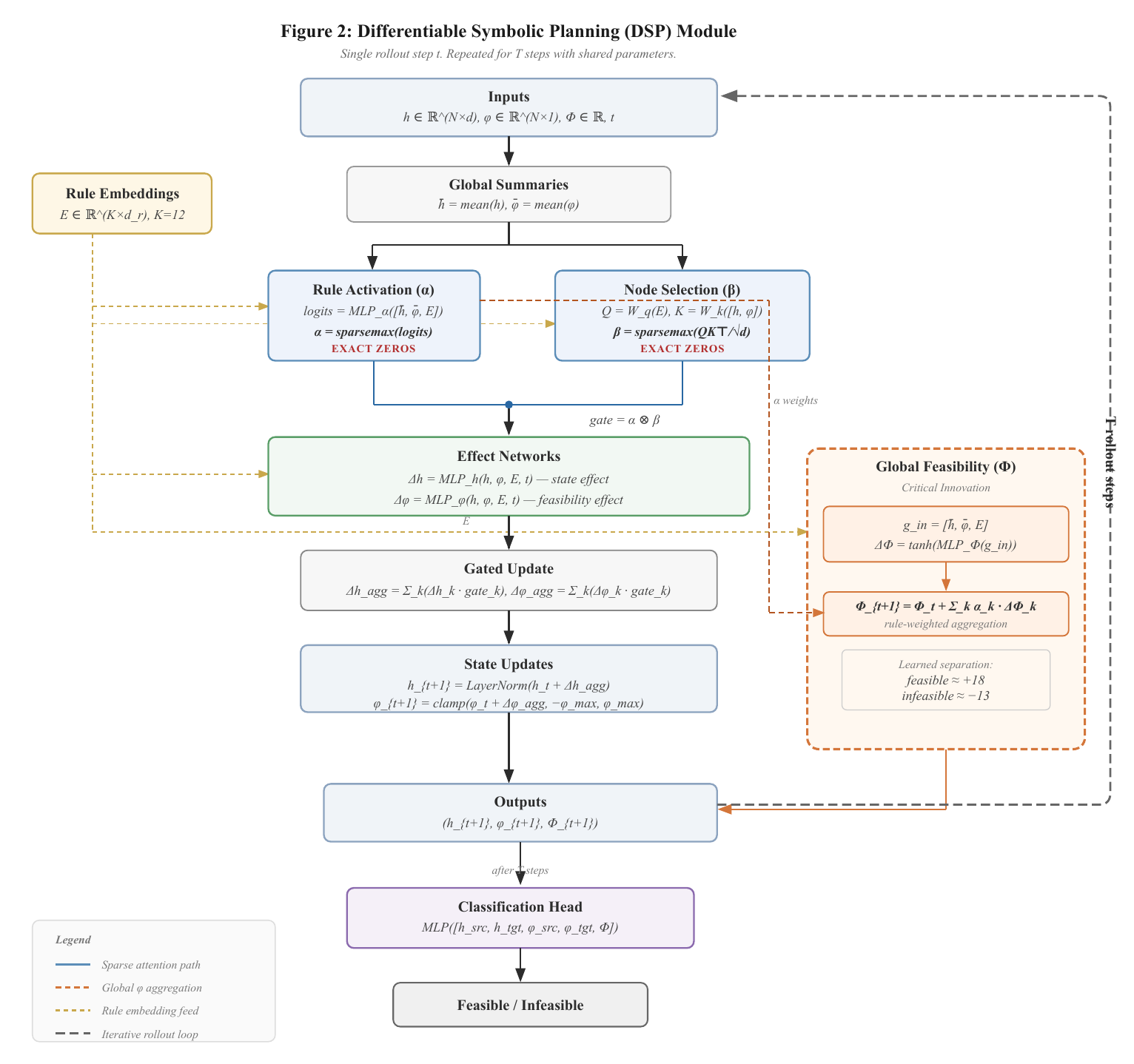}
    \caption{DSP Module detailed architecture. The module computes global summaries, applies sparsemax for discrete rule activation ($\alpha$) and node selection ($\beta$), computes gated effects, and aggregates local feasibility into the critical global $\Phi$ signal. Ablation evidence shows removing global $\phi$ causes 34-point accuracy collapse.}
    \label{fig:dsp_module}
\end{figure}

The DSP module (Figure~\ref{fig:dsp_module}) performs five operations at each rollout step:

\paragraph{Rule Activation ($\alphavec$).}
Given $K$ learnable rule embeddings $\E \in \R^{K \times d_r}$, compute which rules should fire:
\begin{align}
    \bar{\hvec} &= \text{mean}(\hvec) \\
    \bar{\phi} &= \text{mean}(\phivec) \\
    \alphavec &= \text{sparsemax}(\text{MLP}_\alpha([\bar{\hvec}, \bar{\phi}, \E]))
\end{align}
Sparsemax produces exact zeros for non-selected rules, enabling discrete selection.

\paragraph{Node Selection ($\betavec$).}
For each rule, determine which nodes it should affect:
\begin{align}
    \mathbf{Q} &= \mathbf{W}_q \E \\
    \mathbf{K} &= \mathbf{W}_k [\hvec, \phivec] \\
    \betavec &= \text{sparsemax}(\mathbf{Q}\mathbf{K}^\top / \sqrt{d})
\end{align}

\paragraph{Effect Computation.}
Compute how selected rules modify states and feasibility:
\begin{align}
    \Delta\hvec &= \text{MLP}_h([\hvec, \phivec, \E, t]) \\
    \Delta\phivec &= \text{MLP}_\phi([\hvec, \phivec, \E, t])
\end{align}

\paragraph{Gated Update.}
Apply effects through combined rule-node gating:
\begin{align}
    \text{gate} &= \alphavec \otimes \betavec \\
    \hvec &\leftarrow \text{LayerNorm}(\hvec + \sum_k \Delta\hvec_k \cdot \text{gate}_k) \\
    \phivec &\leftarrow \text{clamp}(\phivec + \sum_k \Delta\phivec_k \cdot \text{gate}_k, -\phi_{\max}, \phi_{\max})
\end{align}

\paragraph{Global $\phi$ Aggregation.}
Update the global feasibility signal:
\begin{align}
    \Delta\Phiglobal &= \tanh(\text{MLP}_\Phi([\bar{\hvec}, \bar{\phi}, \E])) \\
    \Phiglobal &\leftarrow \Phiglobal + \sum_k \alpha_k \cdot \Delta\Phiglobal_k
\end{align}

\subsection{Classification}

For tasks with designated source/target nodes (reachability, planning):
\begin{equation}
    \text{logits} = \text{MLP}_{\text{cls}}([\hvec_{\text{src}}, \hvec_{\text{tgt}}, \phi_{\text{src}}, \phi_{\text{tgt}}, \Phiglobal])
\end{equation}

For global classification (SAT):
\begin{equation}
    \text{logits} = \text{MLP}_{\text{cls}}([\text{mean}(\hvec), \text{mean}(\phivec), \Phiglobal])
\end{equation}

\subsection{Sparsemax: Enabling Discrete Differentiable Reasoning}

\begin{figure}[t]
    \centering
    \includegraphics[width=0.8\textwidth]{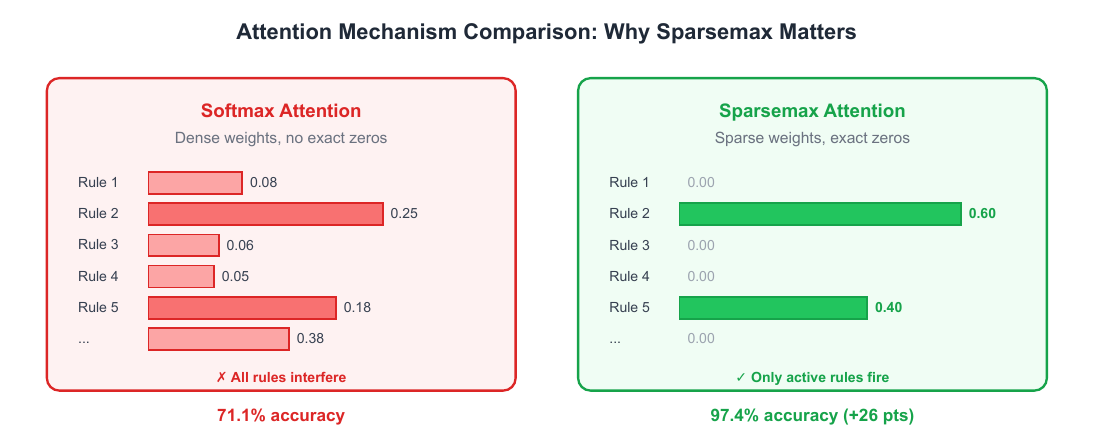}
    \caption{Comparison of softmax vs sparsemax attention. Softmax distributes weight across all rules, causing interference. Sparsemax produces exact zeros, enabling discrete rule selection. This difference accounts for 26 percentage points in accuracy (71.1\% vs 97.4\%).}
    \label{fig:attention}
\end{figure}

Softmax attention assigns nonzero weight to all elements, preventing discrete rule selection. Sparsemax \citep{martins2016softmax} projects onto the probability simplex:
\begin{equation}
    \text{sparsemax}(\mathbf{z}) = \argmin_{\mathbf{p} \in \Delta^{n-1}} \|\mathbf{p} - \mathbf{z}\|^2
\end{equation}
This produces exact zeros for low-scoring elements while remaining differentiable. For constraint reasoning, when only rules 2 and 5 should fire, sparsemax assigns $\alphavec = [0, 0.6, 0, 0, 0.4, 0, \ldots]$ rather than softmax's dense distribution (Figure~\ref{fig:attention}).

\subsection{Why Global $\Phi$ Matters}

Local feasibility values $\phi_i$ capture per-node evidence, but constraint satisfaction is often a global property. Path existence depends on the \emph{entire} graph structure, not individual nodes. The global signal $\Phiglobal$ aggregates local evidence through learned, rule-conditioned combination. Each rule contributes to $\Phiglobal$ based on its activation weight $\alpha_k$ and the global state. This enables the network to learn complex aggregation patterns---some rules may detect local violations while others confirm global connectivity.

Ablation confirms this design: removing global $\phi$ drops accuracy from 98\% to 64\%, a 34-point collapse (Table~\ref{tab:ablation}).

%==============================================================================
\section{Experiments}
%==============================================================================

We evaluate UCK+DSP on three constraint reasoning benchmarks, each requiring different reasoning patterns and exhibiting distinct generalization challenges.

\subsection{Experimental Setup}

\paragraph{Training.} AdamW optimizer with learning rate $3 \times 10^{-4}$, weight decay $10^{-2}$, cosine annealing. Gradient clipping at norm 1.0. 30 epochs. Batch size 32.

\paragraph{Architecture.} $d_{\text{model}} = 64$, $d_{\text{rule}} = 64$, $K = 12$ rules, $T = 4$ rollout steps, $\phi_{\max} = 6.0$, dropout $= 0.1$.

\paragraph{Baselines.}
\begin{itemize}
    \item \textbf{UCK (no DSP)}: Graph attention only, no feasibility channel or rules
    \item \textbf{GIN}: Graph Isomorphism Network \citep{xu2019powerful}, strong GNN baseline
\end{itemize}

\paragraph{Evaluation.} 5 random seeds. Report mean $\pm$ std for overall accuracy, per-class accuracy, and balance score.

\subsection{Results}

\begin{table}[t]
\centering
\caption{Results across three constraint reasoning benchmarks with significant size generalization. UCK+DSP achieves highest accuracy with balanced performance on both positive and negative classes.}
\label{tab:main_results}
\begin{tabular}{@{}llcccc@{}}
\toprule
\textbf{Task} & \textbf{Model} & \textbf{Overall} & \textbf{Positive} & \textbf{Negative} & \textbf{Balance} \\
\midrule
\multirow{3}{*}{\shortstack[l]{Planning\\(8$\times$8 $\to$ 16$\times$16)}} 
& UCK (no DSP) & 59.7\% & 20.1\% & 99.3\% & 0.203 \\
& GIN & 81.1\% & 92.3\% & 69.9\% & 0.757 \\
& \textbf{UCK+DSP} & \textbf{97.4\% $\pm$ 0.1\%} & \textbf{99.9\%} & \textbf{94.8\%} & \textbf{0.949} \\
\midrule
\multirow{2}{*}{\shortstack[l]{SAT\\(10 $\to$ 20 vars)}}
& GIN & 93.8\% $\pm$ 0.5\% & 93.7\% & 93.9\% & 0.999 \\
& \textbf{UCK+DSP} & \textbf{96.4\% $\pm$ 0.7\%} & \textbf{99.6\%} & 93.2\% & 0.936 \\
\midrule
\multirow{3}{*}{\shortstack[l]{Reachability\\(12 $\to$ 30 nodes)}}
& UCK (no DSP) & 73.4\% & 56.2\% & 90.6\% & 0.620 \\
& GIN & 80.5\% & 75.3\% & 85.7\% & 0.879 \\
& \textbf{UCK+DSP} & \textbf{82.7\% $\pm$ 1.2\%} & 77.1\% & \textbf{88.3\%} & 0.873 \\
\bottomrule
\end{tabular}
\end{table}

\paragraph{Planning Feasibility.}
Given a gridworld with obstacles, start position, and goal position, predict whether a feasible path exists. Training on 8$\times$8 grids, testing on 16$\times$16 grids (4$\times$ size increase). UCK+DSP achieves 97.4\% overall accuracy with near-perfect balance (0.949). The ablated UCK collapses to 20\% on feasible cases---exactly the class collapse phenomenon.

\paragraph{Boolean Satisfiability.}
Given a CNF formula as a bipartite graph (variables $\leftrightarrow$ clauses), predict satisfiability. Training on 10 variables, testing on 20 variables (2$\times$ increase). UCK+DSP achieves 96.4\% with statistical significance over GIN ($p = 0.013$).

\paragraph{Graph Reachability.}
Given a directed graph with source and target nodes, predict path existence. Training on 12 nodes, testing on 30 nodes (2.5$\times$ increase). UCK+DSP improves over baselines with balanced performance.

%==============================================================================
\section{Ablation Studies}
%==============================================================================

\subsection{Necessity of $\phi$ Channel}

\begin{table}[t]
\centering
\caption{Ablation study on the planning task. Global $\phi$ aggregation is critical for generalization.}
\label{tab:ablation}
\begin{tabular}{@{}lccc@{}}
\toprule
\textbf{Configuration} & \textbf{8$\times$8 Acc} & \textbf{16$\times$16 Acc} & \textbf{16$\times$16 INFEAS} \\
\midrule
Full DSP & 97.0\% & 98.0\% & 96.0\% \\
No $\phi$ channel & 96.8\% & 58.8\% & 17.6\% \\
No global $\phi$ & 96.6\% & 64.0\% & 28.0\% \\
$\phi$ in keys only & 97.0\% & 98.0\% & 96.0\% \\
$\phi$ in effects only & 96.4\% & 98.2\% & 96.4\% \\
No DSP (baseline) & 82.8\% & 78.4\% & 85.2\% \\
\bottomrule
\end{tabular}
\end{table}

Removing $\phi$ entirely causes INFEASIBLE accuracy to collapse from 96\% to 17.6\%. Removing only global $\phi$ causes collapse to 28\%---global aggregation is critical.

\subsection{Attention Type Comparison}

\begin{table}[t]
\centering
\caption{Comparison of attention mechanisms. Sparsemax is essential for stable, high-accuracy performance.}
\label{tab:attention}
\begin{tabular}{@{}lcc@{}}
\toprule
\textbf{Attention Type} & \textbf{16$\times$16 Accuracy} & \textbf{Variance} \\
\midrule
\textbf{Sparsemax} & \textbf{97.4\%} & \textbf{$\pm$0.1\%} \\
Softmax & 71.1\% & $\pm$2.8\% \\
Entmax ($\alpha$=1.5) & 60.1\% & $\pm$8.3\% \\
\bottomrule
\end{tabular}
\end{table}

Sparsemax achieves 26 points higher accuracy than softmax with 6$\times$ lower variance. Entmax, despite being between softmax and sparsemax, performs worst---the intermediate sparsity regime appears unstable.

\subsection{Learned $\phi$ Semantics}

\begin{table}[t]
\centering
\caption{The global $\phi$ signal learns interpretable semantics without supervision.}
\label{tab:phi_stats}
\begin{tabular}{@{}lcc@{}}
\toprule
\textbf{Class} & \textbf{Global $\phi$ Mean} & \textbf{Global $\phi$ Std} \\
\midrule
FEASIBLE & $+18.0$ & 6.4 \\
INFEASIBLE & $-13.5$ & 15.9 \\
\textbf{Separation} & \textbf{31.5} & --- \\
\bottomrule
\end{tabular}
\end{table}

\begin{figure}[t]
    \centering
    \includegraphics[width=0.75\textwidth]{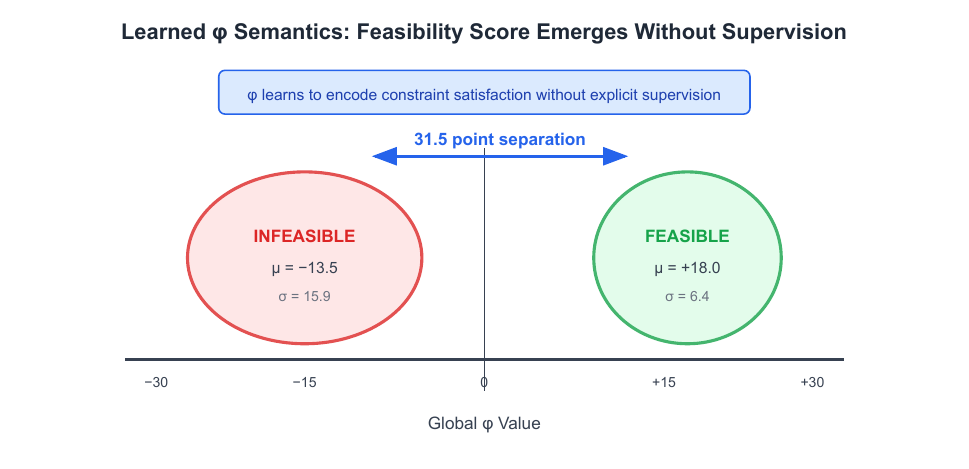}
    \caption{Learned $\phi$ semantics. The global feasibility signal learns to separate feasible ($\mu = +18$) from infeasible ($\mu = -13$) cases with 31.5-point separation, emerging purely from training without explicit supervision of $\phi$ values.}
    \label{fig:phi_semantics}
\end{figure}

Feasible cases produce strongly positive $\phi$; infeasible cases produce negative $\phi$ (Figure~\ref{fig:phi_semantics}). This 31-point separation emerges purely from the classification signal---we never supervise $\phi$ values directly. The magnitude and polarity of the global $\Phi$ signal are task-dependent and emerge through training without explicit supervision; positive and negative values correspond to feasibility semantics only after domain-specific learning.

%==============================================================================
\section{Analysis}
%==============================================================================

\subsection{Why Does Class Collapse Occur?}

Standard GNNs exhibit class collapse under size generalization because:
\begin{enumerate}
    \item \textbf{No persistent state}: Information about constraint satisfaction is lost across layers
    \item \textbf{Dense attention}: All nodes contribute to all updates, preventing focused reasoning
    \item \textbf{Implicit aggregation}: Pooling operations cannot capture complex global constraints
\end{enumerate}
DSP addresses each: $\phi$ provides persistent state, sparsemax enables focused updates, and global $\Phi$ explicitly aggregates constraint evidence.

\subsection{Computational Efficiency}

DSP adds minimal overhead to the base UCK architecture. The full model remains lightweight and practical for deployment, as shown in Table~\ref{tab:efficiency}.

\begin{table}[H]
\centering
\caption{Computational requirements. DSP adds minimal overhead while providing substantial accuracy gains.}
\label{tab:efficiency}
\begin{tabular}{@{}ll@{}}
\toprule
\textbf{Metric} & \textbf{Value} \\
\midrule
Parameters & $\sim$64,000 \\
Training time & $\sim$11 min (T4 GPU) \\
Inference & $<$1ms per sample \\
\bottomrule
\end{tabular}
\end{table}

\subsection{Limitations}

We acknowledge several limitations of the current work:
\begin{itemize}
    \item \textbf{Bounded rollout}: $T=4$ steps may be insufficient for very deep reasoning chains requiring many sequential constraint propagations.
    \item \textbf{Binary classification}: The current formulation handles yes/no feasibility questions, not certificate generation (producing the actual satisfying assignment or path).
    \item \textbf{Graph structure required}: Input must be representable as a graph, which may require domain-specific encoding for some applications.
    \item \textbf{Fixed rule count}: The number of rules $K$ is a hyperparameter; adaptive rule allocation remains future work.
\end{itemize}

%==============================================================================
\section{Conclusion}
%==============================================================================

We introduced Differentiable Symbolic Planning (DSP), a neural architecture for constraint reasoning that maintains explicit feasibility tracking, aggregates local evidence into global signals, and uses sparse attention for discrete rule selection. Integrated into the Universal Cognitive Kernel (UCK), DSP achieves state-of-the-art performance on planning (97.4\%), SAT (96.4\%), and reachability (82.7\%) benchmarks while maintaining balanced accuracy across classes under significant size generalization.

Our ablation studies reveal that global $\phi$ aggregation is critical---removing it causes 34-point accuracy collapse---and that sparsemax attention is essential for stable, high-accuracy performance. The learned $\phi$ signal exhibits interpretable semantics, with feasible/infeasible separation emerging from training without explicit supervision.

DSP demonstrates that neural networks can perform structured constraint reasoning when equipped with appropriate architectural inductive biases. Future work includes extending to multi-step certificate generation, scaling to larger problems, and integrating DSP with planning and verification systems.

%==============================================================================
% References
%==============================================================================

\bibliographystyle{plainnat}
\bibliography{references}

%==============================================================================
% Appendix
%==============================================================================
\newpage
\appendix

\section{Hyperparameters}
\label{app:hyperparams}

\begin{table}[h]
\centering
\begin{tabular}{@{}ll@{}}
\toprule
\textbf{Parameter} & \textbf{Value} \\
\midrule
$d_{\text{model}}$ & 64 \\
$d_{\text{rule}}$ & 64 \\
Number of rules ($K$) & 12 \\
Rollout steps ($T$) & 4 \\
$\phi_{\max}$ & 6.0 \\
Dropout & 0.1 \\
Learning rate & $3 \times 10^{-4}$ \\
Weight decay & $10^{-2}$ \\
Batch size & 32 \\
Epochs & 30 \\
\bottomrule
\end{tabular}
\caption{Hyperparameters used across all experiments.}
\end{table}

\section{Dataset Statistics}
\label{app:datasets}

\begin{table}[h]
\centering
\begin{tabular}{@{}lcccc@{}}
\toprule
\textbf{Dataset} & \textbf{Train Size} & \textbf{Test Size} & \textbf{Train Examples} & \textbf{Test Examples} \\
\midrule
Planning & 8$\times$8 & 16$\times$16 & 10,000 & 2,000 \\
SAT & 10 vars & 20 vars & 10,000 & 2,000 \\
Reachability & 12 nodes & 30 nodes & 10,000 & 2,000 \\
\bottomrule
\end{tabular}
\caption{Dataset statistics for all three benchmarks.}
\end{table}

\section{Sparsemax Implementation}
\label{app:sparsemax}

\begin{algorithm}[h]
\caption{Sparsemax Forward Pass}
\begin{algorithmic}[1]
\REQUIRE Input $\mathbf{z} \in \R^n$
\STATE Sort $\mathbf{z}$ in descending order to get $\mathbf{z}_{(1)} \geq \mathbf{z}_{(2)} \geq \cdots \geq \mathbf{z}_{(n)}$
\STATE Compute cumulative sums $c_k = \sum_{j=1}^{k} \mathbf{z}_{(j)}$
\STATE Find $k^* = \max\{k : 1 + k \cdot \mathbf{z}_{(k)} > c_k\}$
\STATE Compute threshold $\tau = (c_{k^*} - 1) / k^*$
\RETURN $\max(\mathbf{z} - \tau, 0)$
\end{algorithmic}
\end{algorithm}

\section{Related Patent Applications}
\label{app:patents}

The architectural innovations described in this paper are protected by the following U.S. provisional patent applications:

\begin{itemize}
    \item U.S. Provisional Patent Application No.\ 63/978,333, ``Differentiable Symbolic Planning Module with Global Feasibility Aggregation and Sparse Attention for Neural Constraint Reasoning,'' filed February 9, 2026.
    \item U.S. Provisional Patent Application No.\ 63/968,549, ``System and Method for Generating Physically Feasible Response Trajectories Using Temporally-Encoded State Representations,'' filed January 26, 2026.
\end{itemize}

\end{document}